\documentclass{ecai}
\usepackage{graphicx}
\usepackage{latexsym}

\usepackage{amsmath} 
\usepackage{caption}
\usepackage{pgfplots}
\pgfplotsset{compat=newest}
\usepackage{graphicx}
\usepackage{subfigure}
\usepackage{color}
\usepackage{booktabs}
\usepackage{multirow}
\usepackage{arydshln}
\usepackage{amssymb}

\usepackage{algorithm}
\usepackage{algpseudocode}

\begin{document}

\begin{frontmatter}

\title{Create and Find Flatness: Building Flat Training Spaces in Advance for Continual Learning}

\author[A]{\fnms{Wenhang}~\snm{Shi}}
\author[B]{\fnms{Yiren}~\snm{Chen}}
\author[B]{\fnms{Zhe}~\snm{Zhao}}
\author[A]{\fnms{Wei}~\snm{Lu}\thanks{Corresponding Author. Email: lu-wei@ruc.edu.cn}}
\author[B]{\fnms{Kimmo}~\snm{Yan}}
\author[A]{\fnms{Xiaoyong}~\snm{Du}}
\address[A]{School of Information and DEKE, MOE, Renmin University of China}
\address[B]{Tencent AI Lab}

\begin{abstract}
Catastrophic forgetting remains a critical challenge in the field of continual learning, where neural networks struggle to retain prior knowledge while assimilating new information. Most existing studies emphasize mitigating this issue only when encountering new tasks, overlooking the significance of the pre-task phase. Therefore, we shift the attention to the current task learning stage, presenting a novel framework, C\&F (\textit{Create and Find Flatness}), which builds a flat training space for each task in advance. Specifically, during the learning of the current task, our framework adaptively \textbf{creates} a flat region around the minimum in the loss landscape. Subsequently, it \textbf{finds} the parameters' importance to the current task based on their flatness degrees. When adapting the model to a new task, constraints are applied according to the flatness and a flat space is simultaneously prepared for the impending task. We theoretically demonstrate the consistency between the created and found flatness. In this manner, our framework not only accommodates ample parameter space for learning new tasks but also preserves the preceding knowledge of earlier tasks. Experimental results exhibit C\&F's state-of-the-art performance as a standalone continual learning approach and its efficacy as a framework incorporating other methods. Our work is available at \url{https://github.com/Eric8932/Create-and-Find-Flatness}.
\end{abstract}

\end{frontmatter}

\section{Introduction}
As both the size of model parameters and the volume of training data persistently expand, recent Large Language Models (LLMs) have exhibited remarkable generalization abilities across a wide array of tasks \cite{bubeck2023sparks}. Nevertheless, the competences and knowledge of language models remain limited to their training data, and even colossal models struggle in specific situations such as financial scenarios using private data \cite{wu2023bloomberggpt} and recommendation systems using user data \cite{liu2023chatgpt}. The high cost of retraining the model on enlarged data hinders language models to constantly acquire new knowledge. Therefore, Continual Learning (CL), which enables a single model to continually learn and adapt over time, maintains its significance in enhancing the ability of language models.

The major challenge CL faces is the phenomenon of \textit{catastrophic forgetting} \cite{mccloskey1989catastrophic}, where the model loses much of the previous acquired knowledge while adapting to new tasks. To tackle this issue, various approaches have been proposed, encompassing data distribution-based, architecture-based and regularization-based methods. Among these, the regularization-based method is closely related to our work, which inhibits excessive model parameter alternations to maintain proficiency in earlier tasks \cite{aljundi2018memory,kirkpatrick2017overcoming}. However, most approaches emphasize applying constraints during model optimization, neglecting the construction of an appropriate training space for continual learning in advance.
That is, the current task optimization aims to identify the global optimal solution, where the loss landscape is sharp. Consequently, certain model parameters become exceedingly sensitive, implying that minor perturbations can trigger substantial performance decline in previous tasks when adapting to new ones \cite{mirzadeh2020understanding}. Flat-minima based measures, which penalize sharpness and find flat local minima, are employed to mitigate the impact of weight perturbations and enhance single-task model generalization. However, they are not directly applicable to task sequences in CL \cite{wen2018smoothout}.

Hence, we propose a novel framework, C\&F (Create and Find), to build flat training spaces in advance for continual learning. Our framework leverages the geometrical properties of flat minima in both old and new task optimization processes. While training on the current task, we create a flat region around the minimum in the loss landscape. The surrounding of low-loss-value points makes the model robust to a specific range of perturbations from new tasks. Moreover, as different parameters' variations yield disparate effects on the loss, we find the unique flatness of each parameter in the current task using Fisher Information \cite{tu2016ranking}. During the model's adaptation to new tasks, we restrict parameter changes within the flat region and regularize them according to distinct flatness degrees. Concurrently, we prepare the flat space for the subsequent task optimization. In the flat training spaces, the model can assimilate new concepts from new data without forgetting previously acquired knowledge.

Our main contributions encompass the following aspects: 1.Our work pioneers the utilization of flatness in crafting apt training spaces in advance for new task optimization, and we theoretically demonstrate the consistency of the flatness in the learning spaces. 2.We propose C\&F, an innovative framework for continual learning, offering flexibility and seamless integration with other advanced CL techniques. 3.We conduct experiments to evaluate C\&F as both a standalone method and a framework incorporating other methods. The state-of-the-art results demonstrate C\&F's effectiveness.

\section{Related Work}
\subsection{Continual Learning}
Existing CL approaches could be divided into three categories: 1. Data distribution-based method maintains model's ability by aligning with the distribution of previously encountered tasks. Saving a subset of old task samples and incorporating them into new task training is most direct, referred as \textit{replay} \cite{de2019episodic,lopez2017gradient}. In addition, distillation using both old and new samples is also explored \cite{monaikul2021continual,rebuffi2017icarl}. Some generate pseudo-data of old tasks for replay \cite{shin2017continual, sun2019lamol} or distillation \cite{wang2022few}; 2. Architecture-based method focuses on allocating a subset of the model's parameters for each task, optimizing the model exclusively on those parameters. Some allocate and utilize existing parameters to maintain a fixed model size \cite{fernando2017pathnet,mallya2018packnet} and others freeze all model parameters and introduce additional parameters for each new task \cite{razdaibiedina2023progressive,rusu2016progressive}. However, the former imposes a limit on the model's capacity and the latter requires expanding the storage occupied by the model as new tasks appear; 3. Regularization-based method typically incorporates an auxiliary regularization term into the loss function to impede excessive model weight alterations \cite{aljundi2018memory,kirkpatrick2017overcoming,zenke2017continual}. \cite{li2017learning} constrains the current model's output to resemble the previous model's output when processing new data. \cite{zhao2020maintaining} matches the magnitude of weight vectors corresponding to each task. Owing to the disparate importance of parameters, C\&F implements a soft constraint on model optimization by adjusting the loss function.

\subsection{Continual Pre-training of Large Language Models}
Large language models have demonstrated remarkable capabilities. Enhancing their abilities towards different objectives necessitates supplementary training. Techniques such as specific domain or task data training are employed to refine and enrich model's knowledge in specific areas \cite{qin2022elle, ke2023continual}. Instruction tuning empowers these models with the ability to follow instructions, bolstering their performance in zero/few-shot scenarios \cite{sanh2021multitask, wu2022continued}. Moreover, for these models to adhere to human standards, Reinforcement Learning from Human Feedback (RLHF) is deployed for alignment tuning \cite{menick2022teaching, ouyang2022training}. All these continual pre-training sacrifice models' generic capabilities in favor of specific abilities, resulting in forgetting.

\subsection{Flat-Minima Optimization}
\textit{Flat-minima optimizers} optimize both loss and sharpness to find flat local minima and leverage the surrounding low loss values to enhance generalization and scalability for a single task \cite{chaudhari2019entropy}. \cite{foret2020sharpness} optimizes the corrupted loss to ensure low losses around the minima and \cite{kwon2021asam} introduces the concept of adaptive sharpness to achieve scale-invariant flatness. 
Given that the flat regions are inherently suitable for model's continual leaning on new tasks, \cite{mirzadeh2020understanding} examines the role of local minima's geometric properties for each task in the overall degree of forgetting during continual learning. To the best of our knowledge, the only work related to our framework is F2M \cite{shi2021overcoming}, which identifies a flat local minimum in the base task and employs it in incremental few-shot learning. Our framework extends to more general applications in continual learning with a more efficient and effective framework.

\begin{figure*}[t]
\centering
\includegraphics[scale = 0.27]{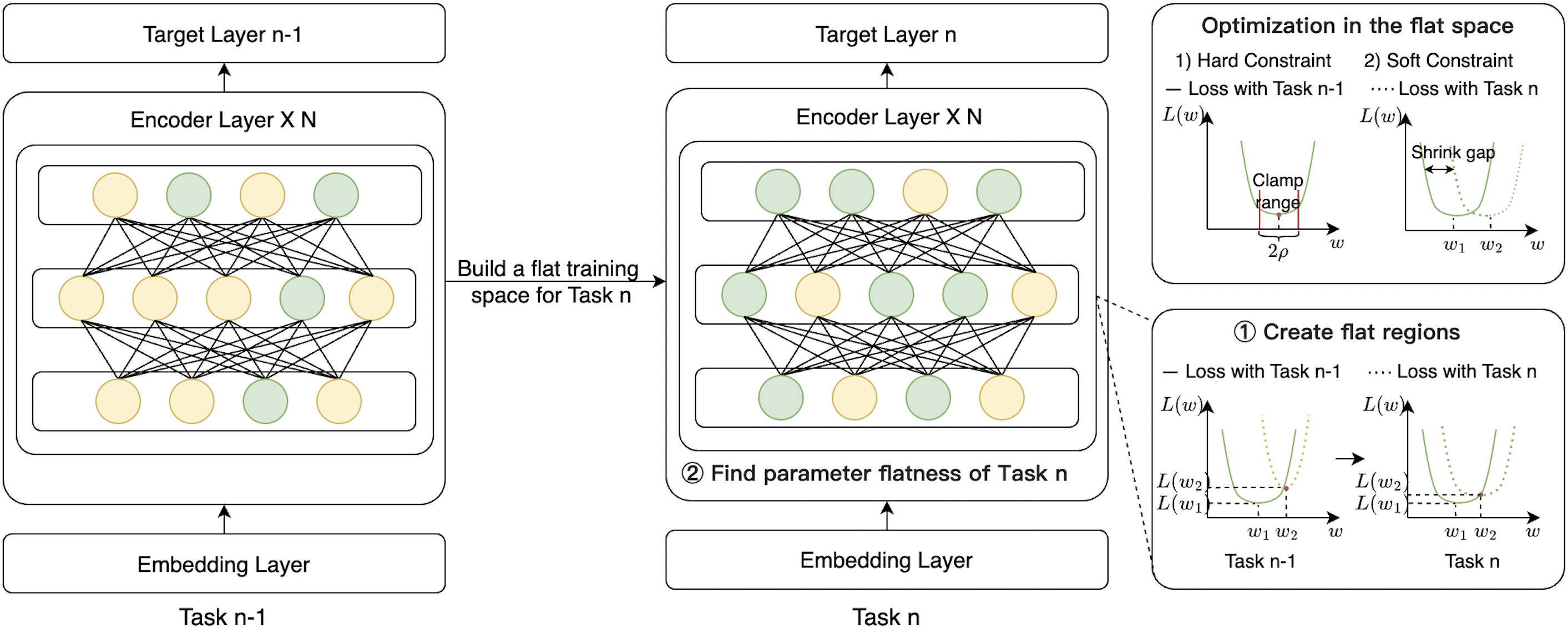}
\caption{Task optimization in the C\&F framework involves two parts: 1.Build a flat training space for the next task: 1) Create a flat region around the minimum and 2) Find parameter flatness for the current task; 2.Optimize the model in the last flat space with two constraints. Yellow and green represent sharpness and flatness, respectively. $L(w_1)$ and $L(w_2)$ denote the loss of task n-1 and n after optimization, respectively.} 
\label{fig_modelpic}
\end{figure*}

\section{Methodology}
In continual learning, the model must adapt to new tasks while preserving proficiency in previous ones. C\&F incorporates flatness characteristics into continual learning process to address the forgetting issue (\S \ref{methodology_meaning}). Specifically, during current task learning, it builds flat training spaces in advance by creating flat regions and finding parameter flatness (\S \ref{methodology_building}).
The new task optimization should be conducted in the last constructed space, while involving preparing a training space for the next task (\S \ref{methodology_optimization}). The whole training process is illustrated in Figure \ref{fig_modelpic}. Last, we justify theoretically the utilization of Fisher Information as flatness indicator by demonstrating its consistency with the Create phase concerning flatness (\S \ref{methodology_consistency}).

\subsection{Meaning of Flatness}\label{methodology_meaning}
Flat local minima means surrounding parameter points all have low loss values. Initially aimed at improving model generalization for a single task, flatness essentially enhances the model's resilience to test set distribution shifts \cite{foret2020sharpness}. Continual learning is a process of encountering perpetually varying task distributions. Therefore, creating flat regions and constraining subsequent tasks to learn within them fosters the model's robustness against continual distribution alterations. In addition, due to the low surrounding loss values around the flat minima, parameter changes do not substantially affect model's performance. Continual learning necessitates ongoing model updates. Learning new tasks within flat regions ensures low losses, i.e., high performances, on prior tasks, addressing the forgetting issue.

\subsection{Build Flat Training Spaces}\label{methodology_building}

\subsubsection{Create Flat Regions}\label{method_Create}
To ensure that the parameter points around minimum all possess low loss values, we optimize for the point with the largest loss value in the entire neighborhood. To formally state the concept, we denote the loss function $\mathcal{L}: \mathcal{W} \times \mathcal{X} \times \mathcal{Y} \to  \mathbb{R_{+} }$ and $\mathcal{S}$ represents the training set. We convert the original $L_\mathcal{S}({\boldsymbol{w}})$ minimization into a MIN-MAX problem, minimizing the maximum loss value around the current point. This approach has been extensively demonstrated, both experimentally and theoretically, to be effective and efficient in creating flat regions \cite{foret2020sharpness,kwon2021asam}. The optimization objective is as follows:
\begin{equation}
\underset{\boldsymbol{w}}{\text{min}} \ \underset{\substack{ \left \| {\boldsymbol{\epsilon}}  \right \|_{2} \le \rho }}{\text{max}} L_\mathcal{S}({\boldsymbol{w}}+{\boldsymbol{\epsilon}}).
\label{EQ1}
\end{equation}
where $\rho$ denotes the range of the flat region.
It is challenging to find the exact optima ${\boldsymbol{\epsilon}}  ^*$ with the maximum loss value, so we employ a first-order approximation. Since $\rho$ determines the learning space for new tasks, we aim to expand it while avoiding the problems associated with its size. To accomplish this, we further adaptively adjust $\rho$ based on the parameter values:
\begin{equation}
\begin{aligned}
\hat{{\boldsymbol{\epsilon}}}({\boldsymbol{w}})=\rho \frac{\mathrm{{\boldsymbol{w}}}^{2} \nabla_{\boldsymbol{w}} L_{S}(\mathbf{{\boldsymbol{w}}})}{\left\|\mathrm{{\boldsymbol{w}}} \nabla_{\boldsymbol{w}} L_{S}(\mathbf{{\boldsymbol{w}}})\right\|_{2}},
\label{EQ2}
\end{aligned}
\end{equation}
where $\mathrm{{\boldsymbol{w}}}^{2}$ is an element-wise operation and $\hat{{\boldsymbol{\epsilon}}}$ is an approximation of ${\boldsymbol{\epsilon}}^*$.  
Finally, we use the gradient of ${\boldsymbol{w}}+\hat{\boldsymbol{\epsilon}}(\boldsymbol{w})$ to update the current parameter point ${\boldsymbol{w}}$ 
\begin{equation}
\begin{aligned}
\left.\nabla_{\boldsymbol{w}} L_{\mathcal{S}}^{Create}(\boldsymbol{w}) \approx \nabla_{\boldsymbol{w}} L_{\mathcal{S}}({\boldsymbol{w}})\right|_{\boldsymbol{w}+\hat{\boldsymbol{\epsilon}}(\boldsymbol{w})},
\label{EQ3}
\end{aligned}
\end{equation}
and the loss for Create is 
\begin{equation}
\begin{aligned}
L_{C}(\boldsymbol{w}) = L_{S}(\boldsymbol{w}+\hat{\boldsymbol{\epsilon}}(\boldsymbol{w}) ).
\label{EQ4}
\end{aligned}
\end{equation}

\subsubsection{Find Parameter Flatness}\label{method_Find}
Within flat regions, sharp parameters still exist. As minor fluctuations in these parameters can result in considerable changes in the loss value, they are crucial to the task and should be found and optimized sparingly. However, determining the exact flatness of each parameter is challenging, so we use Fisher Information as a rapid approximation of flatness \cite{tu2016ranking}. It gauges importance by indicating the extent to which altering the parameter impacts the model's output. Due to the assumption of parameter independence, we only consider the diagonal elements in the Fisher Information matrix. The computation is as follows:
\begin{equation}
\begin{aligned}
F\left({\boldsymbol{w}}_{{\boldsymbol{i}}}\right)=\frac{1}{|S|} \sum_{j=1}^{|S|}\left(\frac{\partial \log p\left(y_{j} \mid  x_{j} ; {\boldsymbol{w}} \right)}{\partial {\boldsymbol{w}}_{{\boldsymbol{i}}}}\right)^{2}.
\label{EQ5}
\end{aligned}
\end{equation}
A smaller Fisher value suggests that the parameters are more likely situated in a flatter region, thereby permitting the model to update more substantially on them.

To enable Fisher Information to represent the flatness of each parameter across all seen tasks, it is computed for every task and accumulated using the following formula:
\begin{equation}
\begin{aligned}
F\left({\boldsymbol{w}}_{{\boldsymbol{i}}}\right)^{'}= \gamma*F\left({\boldsymbol{w}}_{{\boldsymbol{i}}}\right)^{'}+F\left({\boldsymbol{w}}_{{{\boldsymbol{i}}}}\right).
\label{EQ6}
\end{aligned}
\end{equation}

\subsection{Optimization in the Flat Spaces}\label{methodology_optimization}
The last task's Creating and Finding flatness constructs a learning space for the current task. Optimizing parameters in it involves two corresponding constraints. 

\textbf{Hard Constraint.} One is a hard constraint for \textbf{Create}, restricting the model to be optimized in the flat region to preserve its performance on the previous task. After each batch optimization, we clamp parameter changes to ensure they remain in the region:
\begin{equation}
\begin{aligned}
\boldsymbol{w}^{\boldsymbol{\star}}-\rho*(\boldsymbol{w}^{\boldsymbol{\star}})^{abs}    \preceq \boldsymbol{w} \preceq \boldsymbol{w}^{\boldsymbol{\star}}+\rho*(\boldsymbol{w}^{\boldsymbol{\star}})^{abs}.
\label{EQ7}
\end{aligned}
\end{equation}
since we create the flat regions adaptively, we also implement an adaptive clamping strategy based on the size of the parameters.

\textbf{Soft Constraint.} We apply a soft constraint for \textbf{Find} step to  incorporate the found flatness into the optimization process. Specifically, we add an L2 loss term to the loss function, with the flatness serving as a unique coefficient for each parameter, as described in \cite{kirkpatrick2017overcoming}.
Due to the smaller flat range of sharp parameters, restricting their update not only sustains performance on old tasks, but also shrinks the gap between successively created regions so that most parameters stay within all regions. The loss for Find is

\begin{equation}
\begin{aligned}
L_{F}({\boldsymbol{w}})=\sum_{{\boldsymbol{i}}} F\left({\boldsymbol{w}}_{{\boldsymbol{i}}}\right)^{'}\left({\boldsymbol{w}}_{i}-{\boldsymbol{w}}_{{\boldsymbol{i}}}^{{{\boldsymbol{\star}}}}\right)^{2}.
\label{EQ8}
\end{aligned}
\end{equation}
where $\boldsymbol{w}^{\boldsymbol{\star}}$ and $\boldsymbol{w}$ denote the optimal parameter vector learned in the last task and the current parameter vector, respectively.

Apart from two constraints from Create and Find, current task's optimization involves the same creating and finding process to prepare a flat space for the next task. The overall optimization loss is
\begin{equation}
\begin{aligned}
L_{O} = L_{C}+\lambda{L_{F}}.
\label{EQ9}
\end{aligned}
\end{equation}
where $\lambda$ is the coefficient for the soft loss. The full C\&F training process is shown in Algorithm1.

\begin{algorithm}
\caption{The training algorithm for C\&F framework}
\begin{algorithmic}[1]
\Require \parbox[t]{\linewidth}{
Loss function $\mathcal{L}: \mathcal{W} \times \mathcal{X} \times \mathcal{Y} \to  \mathbb{R_{+} }$,

Train set $\mathcal{S} \triangleq \cup_{i=1}^{n}\left\{\left(\boldsymbol{x}_{i}, \boldsymbol{y}_{i}\right)\right\}$, Batch size b, Learning rate $\alpha$,

Flat region range $\rho$, Coefficient $\lambda$ $\gamma$, Weight  $\mathbf{w}_{0}$, Task $t\ge1$. }

\Ensure {Trained weight $\mathbf{w}$.}

\State {\textcolor{blue}{// Optimization for Task 1 and Build a Flat Space for Task 2}}

\State Initialize weight $\mathbf{w}_{1}:=\mathbf{w}_{0}$.
\State \textbf{Create} a flat region around minimum


\While{not converged}
    \State Sample a mini-batch $\mathcal{B}$ of size b from $\mathcal{S}$,
    \State Compute $\hat{\boldsymbol{\epsilon}}(\boldsymbol{w})$ with Eq.2,
    \State Compute loss for \textbf{Create} objective with Eq.3,
    \State Compute gradient $\boldsymbol{g}_{C}$,
    \State Update weights: $\boldsymbol{w}_{1}=\boldsymbol{w}_{1}-\alpha \boldsymbol{g}_{C}$.
\EndWhile

\State \textbf{Find} flatness $F(\boldsymbol{w})$ for each parameter in $\boldsymbol{w}_{1}$ with Eq.5.

\For{task $t = 2, \dots, M$}

    \State{\textcolor{blue}{// Optimization for Task t and Build a Space for Task t+1}}
    \State Initialize weight $\mathbf{w}_{t}:=\mathbf{w}_{t-1}$, $\mathbf{w}^{\boldsymbol{\star}}:=\mathbf{w}_{t-1}$.
    \While{not converged}
        \State Sample a mini-batch $\mathcal{B}$ of size b from $\mathcal{S}$,
        \State Compute gradient $\boldsymbol{g}_{C}$ for \textbf{Create} objective,
        \State Compute gradient $\boldsymbol{g}_{F}$ for \textbf{Find} flatness with Eq.8,
        \State Update weights: $\boldsymbol{w}_{t}=\boldsymbol{w}_{t}-\alpha (\boldsymbol{g}_{C} + \lambda\boldsymbol{g}_{F})$,
        \State Clamp the parameters $\boldsymbol{w}_{t}$ with Eq.7.
    \EndWhile
    \State \textbf{Find} flatness $F(\boldsymbol{w})$ of $\boldsymbol{w}_{t}$ with Eq.5.
    \State Accumulate flatness $F(\boldsymbol{w})^{'}$ with Eq.6.
\EndFor
\State return $\boldsymbol{w}_{t}$;
\end{algorithmic}
\end{algorithm}

\subsection{Flatness Consistency between Create and Find}\label{methodology_consistency}
The selection of Fisher Information as the flatness indicator is due to its consistency with Create concerning flatness. To prove this, we illustrate that the process of creating also minimizes the Fisher values of parameters.
Firstly, min-maximization of loss optimizes both the task loss and the sharpness. 
According to Eq.\ref{EQ1}:
\begin{eqnarray}
\begin{aligned}
L_{\mathcal{S}}^{Create}(\boldsymbol{w}) &= \underset{\left \| {\boldsymbol{\epsilon}}  \right \|_{2} \le \rho}{\max} L_\mathcal{S}({\boldsymbol{w}}+{\boldsymbol{\epsilon}}) \\
&= (\underset{\left \| {\boldsymbol{\epsilon}}  \right \|_{2} \le \rho}{\max} L_\mathcal{S}({\boldsymbol{w}}+{\boldsymbol{\epsilon}}) - L_\mathcal{S}({\boldsymbol{w}}))+L_\mathcal{S}({\boldsymbol{w}}).
\label{EQ10}
\end{aligned}
\end{eqnarray}

Since ${\boldsymbol{\epsilon}}$ is around 0, the sharpness could be defined:
\begin{equation}
\begin{aligned}
sharpness=\max_{{\left \| {\boldsymbol{\epsilon}}  \right \|_{2} \le \rho }} L_\mathcal{S}({\boldsymbol{w}}+{\boldsymbol{\epsilon}}) - L_\mathcal{S}({\boldsymbol{w}}),
\end{aligned}
\label{EQ11}
\end{equation}
Substituting it back, we then have:
\begin{equation}
\begin{aligned}
L_{\mathcal{S}}^{Create}(\boldsymbol{w}) =
sharpness+L_\mathcal{S}({\boldsymbol{w}}).
\label{EQ12}
\end{aligned}
\end{equation}
Therefore, in the process of create, the task loss and the sharpness are optimized simultaneously. 

Then we prove that minimizing the sharpness equals to minimizing the Fisher Info. According to Eq.\ref{EQ2}, the approximately optimal ${\boldsymbol{\epsilon}}^{*}({\boldsymbol{w}})$ is:
\begin{equation}
\begin{aligned}
{\boldsymbol{\epsilon}}^{*}({\boldsymbol{w}})
\approx \rho \frac{ \nabla_{\boldsymbol{w}} L_{S}(\mathbf{{\boldsymbol{w}}})}{\left\| \nabla_{\boldsymbol{w}} L_{S}(\mathbf{{\boldsymbol{w}}})\right\|_{2}},
\label{EQ13}
\end{aligned}
\end{equation}
Notice that we don't consider adaptively adjusting $\rho$ here, so it is different from Eq.\ref{EQ2}. Substituting it back to Eq.\ref{EQ11} and approximating the sharpness via a first-order Taylor expansion of $L_\mathcal{S}({\boldsymbol{w}}+\hat{{\boldsymbol{\epsilon}}}({\boldsymbol{w}}))$ w.r.t ${\boldsymbol{\epsilon}}$ around 0, obtaining:
\begin{equation}
\begin{aligned}
sharpness &\approx  {\boldsymbol{\epsilon}}^{*}({\boldsymbol{w}})^{T}\nabla_{\boldsymbol{w}} L_{S}(\mathbf{{\boldsymbol{w}}})\\
&=\rho \left\| \nabla_{\boldsymbol{w}} L_{S}(\mathbf{{\boldsymbol{w}}})\right\|_{2},
\label{EQ14}
\end{aligned}
\end{equation}
Denoting $g({\boldsymbol{w}})=\left\| \nabla_{\boldsymbol{w}} L_{S}(\mathbf{{\boldsymbol{w}}})\right\|_{2}$ as the norm of gradient of the loss function and $I({\boldsymbol{w}})= \mathbb{E}\left [g({\boldsymbol{w}}) {g({\boldsymbol{w}})^{T}}  \right ]$ as Fisher Information matrix.
Now, let's consider the square of the Euclidean norm of the gradient:
\begin{equation}
\begin{aligned}
\left\|g({\boldsymbol{w}})  \right\|^{2}_{2} = g({\boldsymbol{w}})^{T}g({\boldsymbol{w}}),
\label{EQ15}
\end{aligned}
\end{equation}
Taking the expected value of both sides:
\begin{equation}
\begin{aligned}
\mathbb{E}\left [ \left\|g({\boldsymbol{w}}) \right\|^{2}_{2} \right ]=\mathbb{E}\left[g({\boldsymbol{w}})^{T}g({\boldsymbol{w}}) \right ]=Tr(I({\boldsymbol{w}})) .
\label{EQ16}
\end{aligned}
\end{equation}
where Tr(•) denotes the trace of a matrix. This reveals that minimizing sharpness essentially minimizes the trace of the Fisher Information matrix. Since we compute only the diagonal elements of the matrix during the Find phase, the trace is solely interested, which represents the overall flatness for all parameters. Thus, the flatness created and found are consistent.

\section{Experiment}

\subsection{Datasets and Experimental Setup}
We validate our method on NLP tasks. For most of the experiments, we follow MBPA++ \cite{de2019episodic} and IDBR \cite{huang2021continual} to use five text classification datasets and map Amazon and Yelp tasks to the same label space, resulting in 33 classes with 4 task layers in total. Since the original datasets are too large, we randomly sample 2000 examples per class for training and validation sets to build sampled-version datasets. We conduct major experiments on the sampled-setting text classification datasets, except when comparing to state-of-the-art methods because they are all trained and validated on the full setting. 
To exclude the effect of task order and sequence length, we follow IDBR \cite{huang2021continual} and construct seven task sequences of various lengths and orders. Our implementation is based on TencentPretrain \cite{zhao2022tencentpretrain}, a pre-training toolkit extended from UER-py \cite{zhao2019uer}. More specific information about datasets and task sequences is in supplemental material.

\subsection{Methods for Comparison}
We compare our C\&F as an individual approach with the following methods:
\textbf{Seq} \cite{yogatama2019learning}, sequentially finetunes the model at each task, and \textbf{Replay} \cite{huang2021continual}, stores and replays the most representative old exemplars while learning new tasks, which could be seen as the lower bound. \textbf{LwF} \cite{li2017learning}, \textbf{EWC} \cite{kirkpatrick2017overcoming} and \textbf{MAS} \cite{aljundi2018memory} are three regularization-based methods. LwF limits the output of the current model to be consistent with the old model. EWC and MAS constraint parameter optimization based on their effects on loss and model output, respectively. 
\textbf{F2M} \cite{shi2021overcoming} finds a flat local minimum in the first task and constraints the following task to update around it. \textbf{ORT} \cite{farajtabar2020orthogonal} stores the gradients of some samples (size is 20 here) for each task to construct the orthogonal basis and projects gradient direction of the subsequent tasks to be orthogonal to the basis. \textbf{MTL} learns all the tasks simultaneously and is considered an upper-bound of continual learning.

Apart from baseline methods, we also compare with the state-of-the-art methods. \textbf{MBPA++} \cite{de2019episodic} replays the seen exemplars at training time and uses KNN to select the most representative examples for local adaptation at test time.
\textbf{LAMOL} \cite{sun2019lamol} converts all tasks into a Q\&A format and trains a language model by learning the task and generating training samples at the same time. \textbf{IDBR} \cite{huang2021continual} disentangles text representation space into a task generic space and a task specific space. \textbf{ProgressivePrompt} \cite{razdaibiedina2023progressive} learns a new soft prompt for each task and sequentially concatenates it with the previously learned ones, while keeping the base model frozen.

All baseline methods except for \textbf{Seq} are equipped with replay. We follow \cite{huang2021continual} to store the most representative training samples at the end of each task's learning and replay them while learning new tasks with the same frequency. It's noted that the state-of-the-art methods except \textbf{ProgressivePrompt} also incorporate replay, even with a higher replaying frequency. The concrete implementation details could be found in supplemental material.

\subsection{Model}
We employ BERT-base-uncased as the pretrained model to ensure a fair comparison, as it has been utilized in all preivious works. It should be noted that our C\&F framework is versatile and applicable to models of any size. While it might be appealing to investigate whether CL techniques can further enhance the capabilities of recently emergent large language models, no benchmark is effective against LLMs' extensive knowledge, as they are all composed of public datasets. Additionally, the high costs associated with LLMs and the closed-source nature of the most powerful models pose challenges for comparisons on them. Consequently, we use the medium-sized model to validate the efficacy of the continual learning methods on benchmarks and select the most effective one.

\begin{table*}[t]
\begin{center}
{\caption{Summary of baseline methods’ performance and error bar on Sampled Setting datasets. Metric is averaged accuracy after training on the last task. For each order, we conduct significance test against the best reproducible model, and $^*$ means that the improvement is significant at 0.05 significance level.}\label{tab_baseline}}
\begin{tabular}{lllllllll}
\\ \hline
\textbf{Method} & \multicolumn{4}{l}{\textbf{Length-3 Task Sequences}}        & \multicolumn{4}{l}{\textbf{Length-5 Task Sequences}}    \\
\textbf{Order} & \textbf{I}     & \textbf{II} & \textbf{III} & \textbf{AVG} & \textbf{IV} & \textbf{V} & \textbf{VI} & \textbf{AVG} \\ \hline
Seq            & 69.91          & 63.91          & 70.78          & 68.20$(\pm{1.00})$            & 68.38          & 67.21         & 69.87          & 68.49$(\pm{1.08})$            \\ 
Replay         & 72.85          & 70.54          & 73.48          & 72.29$(\pm{0.46})$            & 71.68          & 71.84         & 74.37          & 72.63$(\pm{0.57})$            \\
\hline
MAS \cite{aljundi2018memory}            & 73.45          & 72.99          & 73.61          & 73.35$(\pm{0.38})$            & 73.78           & 75.22         & 74.30          & 74.43$(\pm{0.5})$             \\
EWC \cite{kirkpatrick2017overcoming}            & 74.39          & 74.12          & 74.47          & 74.33$(\pm{0.19})$            & 75.15           & 74.51         & 75.29          & 74.98$(\pm{0.25})$             \\
LwF \cite{li2017learning}            & 75.17          & 74.21          & 74.57          & 74.65$(\pm{0.24})$            & 75.37           & 74.98         & 75.48          & 75.28$(\pm{0.21})$             \\
F2M \cite{shi2021overcoming}            & 73.56          & 71.43          & 73.71          & 72.90$(\pm{0.37})$            & 73.08          & 72.49         & 74.61          & 73.39$(\pm{0.29})$            \\
ORT \cite{farajtabar2020orthogonal} & 73.82 & 71.50&  73.78 & 72.88$(\pm{0.44})$ & 72.58 & 74.79 & 73.03 & 73.42$(\pm{0.67})$ \\
C\&F       & \textbf{75.89}$^*$ & \textbf{74.81}$^*$ & \textbf{75.01}$^*$ & \textbf{75.24}$(\pm{0.19})$   & \textbf{76.61}$^*$ & \textbf{76.46}$^*$ & \textbf{76.06}$^*$ & \textbf{76.38}$(\pm{0.23})$   \\ \hline
MTL            & 75.31           & 75.31          & 75.31          & 75.31$(\pm{0.08})$            & 76.94          & 76.94         & 76.94          & 76.94$(\pm{0.09})$ \\ \hline
\end{tabular}
\end{center}
\end{table*}

\begin{table*} 
\begin{center}
{\caption{Summary of previous state-of-the-art methods’ performances on Full Setting datasets. Metric is averaged accuracy after training on
the last task. Results of other methods are fetched from the original papers, and C\&F's results are averaged over 3 runs.}\label{tab_sota}}

\begin{tabular}{llllllllll}
\hline
\textbf{Method} & \multicolumn{5}{l}{\textbf{Length-5 Task Sequences}}                             \\
\textbf{Order} & \textbf{IV}    & \textbf{V}    & \textbf{VI}    & \textbf{VII}    & \textbf{AVG} \\ \hline
MBPA++ \cite{de2019episodic}        & 70.7          & 70.2          & 70.9          & 70.8          & 70.7             \\
MBPA++ \cite{sun2019lamol}       & 74.9          & 73.1          & 74.9          & 74.1          & 74.3             \\
LAMOL \cite{sun2019lamol}         & 76.1          & 76.1          & 77.2          & 76.7          & 76.5             \\
IDBR \cite{huang2021continual}            & 75.9          & 76.2          & 76.4          & 76.7          & 76.3             \\
ProgressivePrompt \cite{razdaibiedina2023progressive}        & 78.0          & 77.7          & \textbf{77.9}          & 77.9          & 77.9             \\

C\&F       & \textbf{78.5} & \textbf{78.2} & \textbf{77.9} & \textbf{78.8} & \textbf{78.4}    \\ \hline 
\end{tabular}
\end{center}
\end{table*}

\section{Result and Analysis}
When training on the current task, we validate with a frequency of 500 steps using validation sets consist of all seen tasks, and select the model with the best performance at the end of the task training. We report the averaged accuracy for text classification tasks on test sets after training on all tasks. Without special statement, all results are averaged over 5 runs.

\subsection{Main Results}
Firstly, we evaluate C\&F as an individual method against baselines on sampled-setting task sequences of length-3 and length-5. As illustrated in Table \ref{tab_baseline}, C\&F outperforms other methods across all 6 orders and the performance enhancement becomes more pronounced as the task sequence gets longer, indicating a diminished accumulation of forgetting for C\&F. Notably, when compared to flatness-based or gradient-based methods like F2M or ORT, our approach yields superior results, demonstrating that a simple employment of gradient or flatness features is insufficient to address forgetting in NLP task sequences. While the overall performance improvements might not be statistically significant, they are far from negligible. The pretrained model's strong capabilities have raised the performance lower bound, rendering further advancements challenging, and our performance closely approaches the upper bound MTL.

In Table \ref{tab_sota}, we compare with the previous state-of-the-art methods: MBPA++ \cite{de2019episodic}, LAMOL \cite{sun2019lamol}, IDBR \cite{huang2021continual} and ProgressivePrompt \cite{razdaibiedina2023progressive} on full-setting datasets. The results show that C\&F achieves the best performances on 4 different orders. These findings suggest that combining continual learning and geometric properties effectively alleviates catastrophic forgetting, and the flat training spaces constructed in advance indeed aid the model's constant adaptations. 

\subsection{Integration with other methods}
Our proposed C\&F can function not only as a stand-alone approach for continual learning, but also as a framework to integrate other approaches. We incorporate the IDBR \cite{huang2021continual} approach and conduct experiments in the full-dataset setting as in Table \ref{tab_sota}. In addition to IDBR's original regularization, C\&F framework creates flat regions and finds parameter flatness during current task learning, and adds the corresponding constraints in optimization. Results in Table \ref{tab_clsner} show that C\&F brings further performance boost to IDBR. Moreover, we extend the experiments to NER tasks, in which a new entity type is learned at a time and classification head's dimensions are continually expanded. We integrate our framework with the state-of-the-art method ExtendNER \cite{monaikul2021continual}, which employs knowledge distillation to replay previous entities while learning a new one using samples annotated exclusively on the new entity type. Full dataset details of NER tasks are available in the supplemental material. Results in Table \ref{tab_clsner} indicate that C\&F further enhances the performance of the SOTA method on NER tasks. These findings suggest that C\&F framework is versatile and CL techniques can be seamlessly and effectively introduced to improve performance further.

\begin{table*}
\center
{\caption{Comparisons between original method and method in C\&F framework. Metric for Text Classification is averaged accuracy after training on the last task. Metric for NER is the averaged F1 over seen entity types on the test set at each step and scores at each step are averaged over all class orders.}\label{tab_clsner}}
\begin{tabular}{lllllllllll}
\hline
\multicolumn{11}{c}{Text Classification}     
 \\ \hdashline
\textbf{Method}     & \multicolumn{2}{l}{IV}                        & \multicolumn{2}{l}{V}           & \multicolumn{2}{l}{VI}                             & \multicolumn{2}{l}{VII}         & \multicolumn{2}{l}{AVG}         \\
IDBR \cite{huang2021continual}      & \multicolumn{2}{l}{75.9}                      & \multicolumn{2}{l}{76.2}        & \multicolumn{2}{l}{76.4}                           & \multicolumn{2}{l}{76.7}        & \multicolumn{2}{l}{76.3}        \\
+C\&F      & \multicolumn{2}{l}{\textbf{77.4}}                      & \multicolumn{2}{l}{\textbf{77.2}}       & \multicolumn{2}{l}{\textbf{77.4}}                           & \multicolumn{2}{l}{\textbf{77.6}}        & \multicolumn{2}{l}{\textbf{77.4}}        \\ \hline
\multicolumn{11}{c}{NER}                                                                                                                                                                                             \\ \hdashline
\textbf{Method}    & \multicolumn{4}{c}{CoNLL-03}                                      & \multicolumn{6}{c}{OntoNotes-5.0}                                                                   \\
          & Step 1                       & Step 2         & Step 3         & Step 4         & Step 1                            & Step 2         & Step 3         & Step 4         & Step 5         & Step 6         \\ \cmidrule(r){2-5} \cmidrule(r){6-11}
ExtendNER \cite{monaikul2021continual} & 86.41                        & 86.47          & 86.41          & 86.41          & 79.80                             & 80.59          & 80.88          & 80.80          & 80.27          & 79.51          \\
+C\&F      & \textbf{87.10}               & \textbf{87.20} & \textbf{87.23} & \textbf{86.98} & \textbf{81.24}                    & \textbf{81.58} & \textbf{81.68} & \textbf{81.41} & \textbf{80.98} & \textbf{80.21} \\ \hline
\end{tabular}
\center
\end{table*}

\subsection{Analysis of CREATE}
To investigate whether the created flatness helps mitigate catastrophic forgetting and provides ample learning space for new tasks, we compare C\&F with F (Find), which discards Create, in terms of learning ability, forgetting and accuracy. Despite the absence of Create, F still retains the parameter clamping after each update for a fair comparison.
We adopt the measure of intransigence, representing the inverse of learning ability, and forgetting from \cite{chaudhry2018riemannian}. We detail the measures in supplemental material. 
Figure \ref{fig_intforacc} displays the comparison of the two methods under sampled-setting order 4 as $\rho$ rises and the updating regions expand. 
C\&F consistently exhibits lower levels of forgetting, suggesting that the creation of flat regions effectively diminishes forgetting. Additionally, C\&F's lower intransigence indicates its enhanced ability to learn new tasks.
Although the improvement in generalization partially stems from the created flatness, it demonstrates that the flat regions do not impede new task learning. It is worth noting that C\&F's intransigence increases when the value of $\rho$ is high, as the flatness comes at the expense of increased loss. Overall, C\&F successfully reduces forgetting and bolsters learning by creating flat regions, as evidenced by the accuracy curves as well.

\begin{figure}[t]
\centerline{\includegraphics[scale = 0.53]{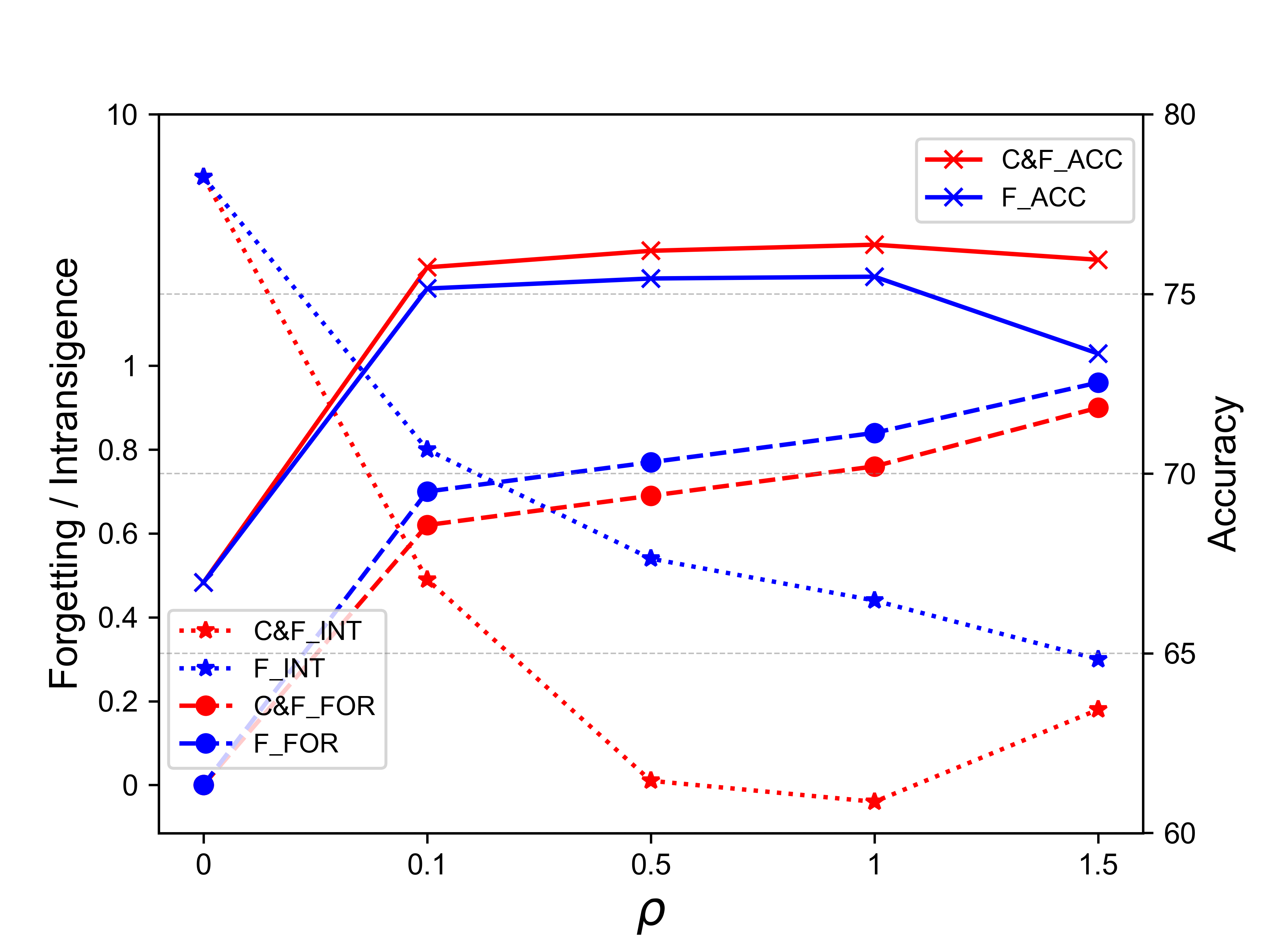}}
\caption{Comparisons of intransigence, forgetting and accuracy between C\&F and F, C\&F without creating, under order 4 as $\rho$ rises. INT is short for intransigence and FOR is short for forgetting. Left y-axis is for intransigence and forgetting, and right is for accuracy.} 
\label{fig_intforacc}
\end{figure}

\subsection{Analysis of FIND}

\subsubsection{Overlap of Regions}\label{find_overlap}
\begin{figure*}[h]
\centering
\begin{tikzpicture}[scale=0.53]
\begin{axis}[
    title = {order4},
    xlabel=task,
    ylabel= sharpness,
    tick align=outside,
    xtick = {0,1,2,3,4},
    ytick = {4,5,6,7,8,9,10},
    ymin=3.7,ymax=10.6,
    legend style={at={(0.13,1.1)},anchor=north, font=\large},
    label style={font=\normalfont},
    ymajorgrids=true,
    grid style=dashed
    ]

\addplot[smooth,mark = *,red] plot coordinates {
    (0,6.1)
    (1,6.85)
    (2,6.94)
    (3,7.64)
    (4,7.28)
};
\addplot[smooth,mark = *,green] plot coordinates {
    (0,6.1)
    (1,7.29)
    (2,8.34)
    (3,8.09)
    (4,8.96)
};
\end{axis}
\end{tikzpicture}
\hspace{0.7em}
\begin{tikzpicture}[scale=0.53]
\begin{axis}[
    title={order5},
    title style={font=\large},
    xlabel=task,
    xlabel style={font=\large},
    tick align=outside,
    xtick = {0,1,2,3,4},
    ytick = {4,5,6,7,8,9,10},
    ymin=3.7,ymax=10.6,
    legend style={at={(0.13,1.1)},anchor=north, font=\large},
    ymajorgrids=true,
    grid style=dashed
    ]

\addplot[smooth,mark = *,red] plot coordinates {
    (0,6.99)
    (1,8.27)
    (2,6.85)
    (3,7.12)
    (4,7.13)
};
\addplot[smooth,mark = *,green] plot coordinates {
    (0,6.99)
    (1,8.45)
    (2,8.61)
    (3,10.13)
    (4,8.51)
};
\end{axis}
\end{tikzpicture}
\hspace{0.7em}
\begin{tikzpicture}[scale=0.53]
\begin{axis}[
    title={order6},
    title style={font=\large},
    xlabel style={font=\large},
    ylabel style={font=\large},
    xlabel=task,
    tick align=outside,
    xtick = {0,1,2,3,4},
    ytick = {4,5,6,7,8,9,10},
    ymin=3.7,ymax=10.6,
    legend style={at={(0.13,1.1)},anchor=north, font=\large},
    ymajorgrids=true,
    grid style=dashed
    ]

\addplot[smooth,mark = *,red] plot coordinates {
    (0,4.58)
    (1,4.70)
    (2,4.10)
    (3,5.78)
    (4,6.18)
};
\addplot[smooth,mark = *,green] plot coordinates {
    (0,4.58)
    (1,4.86)
    (2,4.19)
    (3,6.62)
    (4,6.74)
};
\end{axis}
\end{tikzpicture}
\hspace{0.7em}
\begin{tikzpicture}[scale=0.53]
\begin{axis}[
    title style={font=\large},
    xlabel style={font=\large},
    ylabel style={font=\large},
    title={order7},
    xlabel=task,
    tick align=outside,
    xtick = {0,1,2,3,4},
    ytick = {4,5,6,7,8,9,10},
    ymin=3.7,ymax=10.6,
    legend style={at={(0.13,1.1)},anchor=north, font=\large},
    ymajorgrids=true,
    grid style=dashed
    ]

\addplot[smooth,mark = *,red] plot coordinates {
    (0,7.4)
    (1,7.6)
    (2,7.74)
    (3,7.62)
    (4,8.09)
};
\addplot[smooth,mark = *,green] plot coordinates {
    (0,7.4)
    (1,9.39)
    (2,8.17)
    (3,8.4)
    (4,8.38)
};
\end{axis}
\end{tikzpicture}

\caption{Comparisons of whole model's sharpness for first task during continual learning on different orders. \textbf{Red lines} denote C\&F and \textbf{green lines} denote C, C\&F without Find. Sharpness is calculated by the log value of the largest eigenvalue of the Hessian Spectrum.}
\label{fig_sharpness}
\end{figure*}
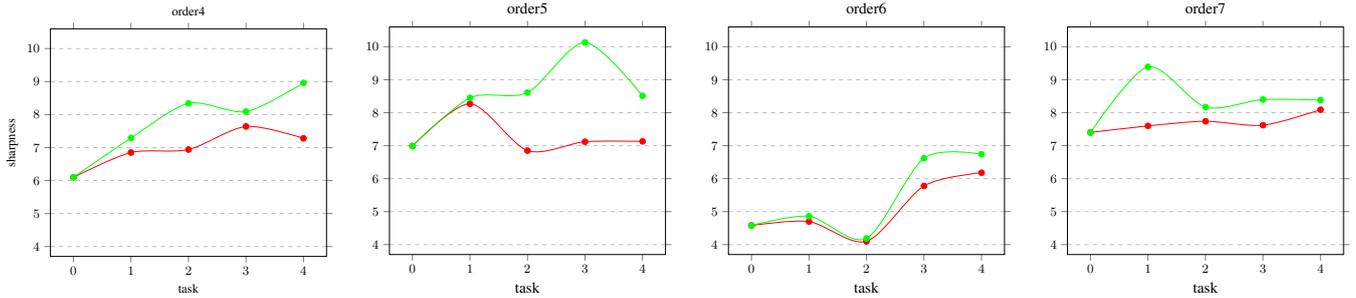

Flat regions are created for each task, but the model only needs updating within the most recent region. Overlapping regions ensure that more parameters remain in areas of low task loss, crucial for maintaining performances across all tasks. Regularizing parameter optimization according to the found flatness helps shrink the gap between the regions.
Experimentally, Figure \ref{fig_sharpness} shows the overall model sharpness change for the first task after trained on each task by C\&F and C (Create), which discards Find and the corresponding loss regularization term. We approximate the Hessian Spectrum using the Lanczos algorithm \cite{ghorbani2019investigation} and choose the largest eigenvalue to characterize sharpness, reporting the log value. As expected, the model trained with Find maintains lower sharpness for the first task as learning new tasks, implying that more parameters reside in the firstly created flat region. Thus, Find enhances the regions' overlap degree, resulting in the construction of superior flat training spaces.

Moreover, although the creation of flat regions flattens all parameters, some parameters remain sharp with a small flat range. Over-optimizing them can cause the model to deviate from the minimum and leave the region. Results in Figure \ref{fig_sharpness} demonstrate that Find further constrains the parameters within regions by preventing excessive optimization on those sharp parameters. It experimentally verifies that the flatness found for each parameter is consistent with the created flatness.

\subsubsection{Flatness Indicator}
We have validated the use of Fisher Information as the flatness indicator in our framework both theoretically (\S \ref{methodology_consistency}) and experimentally (\S \ref{find_overlap}). Furthermore, we experiment with alternative indicators from MAS \cite{aljundi2018memory} and SI \cite{zenke2017continual} and replace Fisher with them in C\&F framework, respectively. We also compare with \textbf{C} (Create), eliminating Find and its corresponding regularization from C\&F, and \textbf{Random}, randomly sampling flatness value for each parameter. Results in Table \ref{tab_find} underscore the superiority of Fisher Information as a flatness indicator within our C\&F framework, proving that it is more conducive to constructing optimal learning spaces.

\begin{table}
\begin{center}
{\caption{Comparisons of different flatness indicator for Find. MAS and SI replace Fisher Information, which is originally used in C\&F, with indicator utilized in MAS \cite{aljundi2018memory} and SI \cite{zenke2017continual}, respectively.}\label{tab_find}}
\begin{tabular}{lllll}
\hline
\textbf{Method}  & \textbf{IV} & \textbf{V} & \textbf{VI} & \textbf{AVG}   \\ \hline
C    & 74.70  & 74.06  & 75.32  & 74.70 \\
Random & 75.34  & 74.67  & 75.49  & 75.17 \\
MAS & 76.26 & 75.03 & 75.87 & 75.72 \\
SI & 76.00 & 75.04 & 75.63 & 75.45 \\
C\&F    & \textbf{76.61} & \textbf{76.46} & \textbf{76.06} & \textbf{76.38} \\ \hline
\end{tabular}
\end{center}
\end{table}

\subsubsection{Sparsity of Find}
Given that continual learning necessitates saving every model for rollback in practical applications, we further investigate the effect of parameter sparsity in Find. The most straightforward approach is to find parameters with the least importance to the last task in each layer and solely optimize on them in the new task. Table \ref{tab_sparseopt} presents the results of varying sparingly optimizing ratios applied to C\&F per layer. The performance gradually increases as the ratio rises, with only 50\% of parameters being optimized achieving performances comparable to full optimization. Sparsity ensures more parameters located in the flat regions, especially those are sharp, which further alleviates catastrophic forgetting. Therefore, C\&F with sharp parameters fixed can be a suitable choice in practice.

\begin{table}
\begin{center}
{\caption{Comparisons of performance with increasing sparingly updating ratio per layer. 100\% is the setting of C\&F.}\label{tab_sparseopt}}
\begin{tabular}{lllll}
\\ \hline
\textbf{SU\_Ratio} & \textbf{IV}     & \textbf{V} & \textbf{VI}     & \textbf{AVG}   \\ \hline
0.1\%                & 69.41          & 72.83      & 68.04          & 70.09 \\ 
1\%                  & 71.26          & 74.77      & 68.60          & 71.54 \\
10\%                  & 74.74          & 75.82      & 73.96          & 74.84 \\
20\%                  & 75.34 & 75.89      & 74.90          & 75.38 \\
50\%                  & 76.26 & 75.92      & 75.83          & 76.00 \\ 
100\%                & \textbf{76.61} & \textbf{76.46} & \textbf{76.06} & \textbf{76.38} \\
 \hline
\end{tabular}
\end{center}
\end{table}

\subsection{Time Comparison}
\begin{figure}[t]
\pgfplotsset{width=9cm, height=4.5cm} 
\centering
	\begin{tikzpicture}
		\begin{axis}
			[ybar, ymin=0,ymax=20,ylabel=Time(hours),
			 symbolic x coords={Seq, Replay, MAS, EWC, LwF, F2M, ORT, C\&F},
			 axis y line=left,
			 axis x line=bottom,
			 nodes near coords,
			enlarge x limits=0.05,] 
			\addplot[draw=yellow, fill=orange] coordinates {(Seq, 0.96) (Replay, 1.55) (MAS, 2.38) (EWC, 1.79) (LwF, 2.40) (F2M, 5.00) (ORT, 19.67) (C\&F, 2.23)}; 
		\end{axis} 
	\end{tikzpicture} 
 \caption{Time comparisons in hours for the main methods on sampled-setting order4 task sequence.}
 \label{fig_time}
 \end{figure}
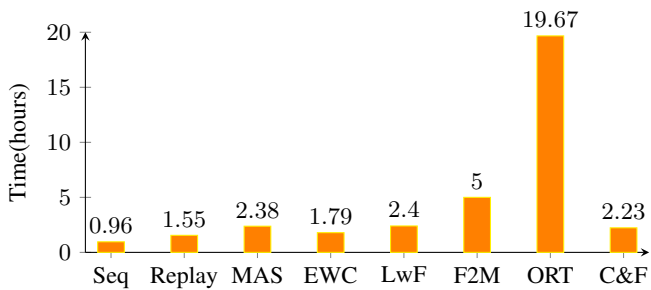

While the C\&F framework offers a distinctive approach that encompasses both the creating and finding flatness processes, it tends to exhibit marginally lower efficiency compared to direct optimization methods. This primarily results from the additional time required for creating flat regions, which may impede the optimization speed. We provide time comparisons in Fig \ref{fig_time}, illustrating the learning time in hours for the main methods on the sampled-setting order4 task sequence, as used in Table \ref{tab_baseline}. The training settings of these methods are consistent. It's noted that C\&F's time extension is insignificant and its training time is considerably less than that of other flatness-based or gradient-based methods like F2M or ORT. This reveals that directly incorporating gradient or flatness characteristics into continual learning may lead to substantial efficiency degradation. 

\subsection{Effect of C\&F Components}

We have devised a generic framework for continual learning, with the core of creating and finding flatness and the associated optimization constraints. We outline each component of C\&F below and examine the effects of ablating them one by one. Firstly, We replay samples saved from previous tasks, which are picked out as the most representative using K-Means 
\cite{macqueen1967classification}. 
In addition, we create a flat region around the minimum of each task. To further mitigate catastrophic forgetting, we incorporate an L2 loss term in optimization, employing random coefficient as the flatness value for each parameter. Furthermore, we find each parameters' flatness and use it as the coefficient of the L2 term specific to each parameter. Finally, we clamp all parameters' optimization within the created regions. As shown in Table \ref{tab_discard}, incrementally discarding each ingredient leads to a decline in accuracy. Thus, every component in the C\&F framework is essential for achieving the optimal performance.

\begin{table}
\begin{center}
{\caption{Effects of ablating different components in C\&F.}\label{tab_discard}}
\begin{tabular}{lllll}
\hline
\textbf{Method}    & \textbf{IV}    & \textbf{V}     & \textbf{VI}    & \textbf{AVG}   \\ \hline
C\&F      & \textbf{76.61} & \textbf{76.46} & \textbf{76.06} & \textbf{76.38} \\ 
- Clamp  & 76.32          & 75.26 & 75.80          & 75.79 \\
- Find   & 75.77 & 73.73 & 75.61 & 75.04 \\
- L2    & 74.10 & 73.64 & 75.21 & 74.31 \\
- Create & 71.68 & 71.84 & 74.37 & 72.63 \\
- Replay & 68.38 & 67.21 & 69.87 & 68.49 \\ \hline
\end{tabular}
\end{center}
\end{table}

\section{Conclusion}
In this work, we introduce C\&F, an innovative framework incorporating geometrical properties of flatness to build flat training spaces in advance for continual learning. C\&F \textbf{creates} a flat region around the minimum and \textbf{finds} each parameters' flatness both with respect to the current task. During the new task's optimization, C\&F constraints parameter changes in the created flat regions and regularizes them according to the found flatness, simultaneously preparing a new flat training space for the subsequent task. As an individual continual learning method, C\&F outperforms the previous state-of-the-art approaches in text classification task sequences. Moreover, when other methods are integrated into C\&F framework, their performances attain further boost. In future research, we aim to explore the flatness's applicability to other fields concerning continual learning.

\ack The work was supported by the National Natural Science Foundation of China under Grant No. 62072458.

\bibliography{ecai}

\appendix
\newpage 

\begin{center}
\Large\textbf{Supplemental Material}
\end{center}

This supplemental material includes more details about some aspects of the main paper. Specifically, it provides concrete datasets' information and task orders for text classification tasks (\S \ref{app_clsdataset}) and NER tasks (\S \ref{app_nerdataset}). Moreover, the specific implementation details employed during the model's continual training and the hyper-parameters for the replay setting are provided (\S \ref{app_implementation}). Last, it expounds upon the Intransigence and Forgetting metrics, which are utilized, which are used to meticulously evaluate the model's learning capability and knowledge change throughout the continual learning process (\S \ref{app_intfor}).

\textit{Notice that the bibliographic references in this supplemental paper are about the bibliography of the main paper.}

\section{Datasets and Task Orders for Text Classification Task}\label{app_clsdataset}
For text classification task, we follow MBPA++ [5] and IDBR [10] to choose five most used datasets, including AG News, Yelp, DBPedia, Amazon and Yahoo! Answer. Details of the five datasets are listed in Table 7. Based on these 5 tasks, we construct 7 different task sequences, including length of 3 and 5. The specific task orders of the sequences are in Table 8:

\begin{table}
\begin{center}
{\caption{Dataset statistics we used for Sampled Setting in classification tasks. Type denotes the domain for task classification. The size of the validation set is the same as the size of the training set.}}\label{tab_clsdatasetssta}
\begin{tabular}{lllll}
\\ \hline
\textbf{Dataset} & \textbf{Class} & \textbf{Type} & \textbf{Train} & \textbf{Test} \\ \hline
AGNews           & 4              &News        & 8000           & 7600          \\
Yelp             & 5              &Sentiment    & 10000          & 7600          \\
Amazon           & 4              &Sentiment    & 10000          & 7600          \\
DBPedia          & 14             &Wikipedia   & 28000          & 7600          \\
Yahoo            & 10             &Q\&A      & 20000          & 7600         \\ \hline
\end{tabular}
\end{center}
\end{table}

\begin{table}
\begin{center}
{\caption{Seven different task sequences used for experiments for classification tasks. The first 6 are used in Sampled Setting. The last 4 are used in Full Setting.}}\label{tab_clsorder}
\begin{tabular}{ll}
\\ \hline
\textbf{Order} & \textbf{Task Sequence}       \\ \hline
I              & ag$\rightarrow$yelp$\rightarrow$yahoo                \\
II              & yelp$\rightarrow$yahoo$\rightarrow$ag                \\
III              & yahoo$\rightarrow$ag$\rightarrow$yelp                \\
IV              & ag$\rightarrow$yelp$\rightarrow$amazon$\rightarrow$yahoo$\rightarrow$dbpedia \\
V              & yelp$\rightarrow$yahoo$\rightarrow$amazon$\rightarrow$dbpedia$\rightarrow$ag \\
VI              & dbpedia$\rightarrow$yahoo$\rightarrow$ag$\rightarrow$amazon$\rightarrow$yelp \\
VII              & yelp$\rightarrow$ag$\rightarrow$dbpedia$\rightarrow$amazon$\rightarrow$yahoo \\ \hline
\end{tabular}
\end{center}
\end{table}

\vspace*{\fill}
\section{Datasets and Class Orders for NER Task}\label{app_nerdataset}
For NER task, we follow [22] to choose two classic datasets: CoNLL-03 English NER and OntoNotes-5.0 English. To ensure enough examples for training, we select named entity types. For CoNLL-03 English, we choose 4 types: Person (PER), Location (LOC), Organization (ORG), and Miscellaneous (MISC). For OntoNotes-5.0 English, we choose 6 types: Organization (ORG), Person (PER), Geo-Political Entity (GPE), Date (DATE), Cardinal (CARD), Nationalities and Religious Political Group (NORP). Dataset statistics are shown in Table 9.

\begin{table*}
\begin{center}
{\caption{Distribution of entity labels in CoNLL-03 and OntoNotes-5.0.}}\label{tab_nerdatasetsta}
\begin{tabular}{lcccclcccccc}
\hline
      & \multicolumn{4}{c}{\textbf{CoNLL-03}} &  & \multicolumn{6}{c}{\textbf{OntoNotes-5.0}}         \\ 
      & PER     & LOC     & ORG     & MISC    &  & ORG    & PER    & GPE    & DATE   & CARD   & NORP  \\ \cline{2-5} \cline{7-12}
Train & 6,532   & 7,125   & 6,271   & 3,398   &  & 24,163 & 22,035 & 21,938 & 18,791 & 10,901 & 9,341 \\
Dev   & 1,829   & 1,832   & 1,325   & 916     &  & 3,798  & 3,163  & 3,649  & 3,208  & 1,720  & 1,277 \\
Test  & 1,597   & 1,664   & 1,654   & 698     &  & 2,002  & 2,134  & 2,546  & 1,787  & 1,005  & 990  \\ \hline
\end{tabular}
\end{center}
\end{table*}

Following [22], apart from selecting entity types, to construct datasets for class-incremental learning, we divide the train/dev sets of the two datasets into 4 and 6 disjoint subsets, $D^1$, $D^1$, ..., respectively: each $D^i$ is annotated only for the entity type $e^i$. And the test set up to step i is annotated for the entity type {$e^1$, ..., $e^i$}. As in task-incemental setting, we also sample 8 and 6 different class/type orders for CoNLL-03 and OntoNotes-5.0, respectively, as shown in Table 10.

\begin{table*}
\begin{center}
{\caption{Different class orders in which entity type are added to the models for each dataset.}}\label{tab_nerorder}
\begin{tabular}{cccc}
\hline
Order & \textbf{CoNLL-03}                                             &  & \textbf{OntoNotes-5.0}                                                                          \\ \cline{1-2} \cline{3-4}
I     & PER$\rightarrow$LOC$\rightarrow$ORG$\rightarrow$MISC &  & ORG$\rightarrow$PER$\rightarrow$GPE$\rightarrow$DATE$\rightarrow$CARD$\rightarrow$NORP \\
II    & PER$\rightarrow$MISC$\rightarrow$LOC$\rightarrow$ORG &  & DATE$\rightarrow$NORP$\rightarrow$PER$\rightarrow$CARD$\rightarrow$ORG$\rightarrow$GPE \\
III   & LOC$\rightarrow$PER$\rightarrow$ORG$\rightarrow$MSC  &  & GPE$\rightarrow$CARD$\rightarrow$ORG$\rightarrow$NORP$\rightarrow$DATE$\rightarrow$PER \\
IV    & LOC$\rightarrow$ORG$\rightarrow$MISC$\rightarrow$PER &  & NORP$\rightarrow$ORG$\rightarrow$DATE$\rightarrow$PER$\rightarrow$GPE$\rightarrow$CARD \\
V     & ORG$\rightarrow$LOC$\rightarrow$MISC$\rightarrow$PER &  & CARD$\rightarrow$GPE$\rightarrow$NORP$\rightarrow$ORG$\rightarrow$PER$\rightarrow$DATE \\
VI    & ORG$\rightarrow$MISC$\rightarrow$PER$\rightarrow$LOC &  & PER$\rightarrow$DATE$\rightarrow$CARD$\rightarrow$GPE$\rightarrow$NORP$\rightarrow$ORG \\
VII   & MISC$\rightarrow$PER$\rightarrow$LOC$\rightarrow$ORG &  &                                                                                        \\
VIII  & MISC$\rightarrow$ORG$\rightarrow$PER$\rightarrow$LOC &  &           \\ \hline                                                                          
\end{tabular}
\end{center}
\end{table*}

\section{Implementation Details}\label{app_implementation}
We use consistent experimental settings to ensure fair comparisons. We set 8 for batch size, 256 for maximum sequence length, 3e-5 for learning rate and use single A100 GPU for training. We use AdamW as optimizer and constant schedule with warm up as scheduler and weight decay is 0.01. We set $\rho$, which decides the size of flat regions, to 0.65. The coefficient $\lambda$ for Find loss, soft constraint, is 50000 and the coefficient $\gamma$ for accumulating Fisher is 0.95. The number of samples used to calculate Fisher Information for each task is 128. All optimizing constraints are for the encoder and the task layer of the old tasks. All hyper-parameters are searched on the sampled version of the datasets. 

For replay, we set store ratio as 0.01 and replay frequency as 20, meaning that current task replay every 20 batch updates. All baseline methods except for \textbf{Seq} are equipped with replay and their replay frequencies are identical. Even more, state-of-the-art techniques, with the exception of ProgressivePrompt [25], leverage replay mechanisms, and some even adopt a more frequent replay schedule to reinforce learning..

\section{Intransigence and Forgetting}\label{app_intfor}
We follow [4] to measure Intransigence $\mathcal{I}_{k}$ after trained on task k as follows:
\begin{equation}
\begin{aligned}
\mathcal{I}_{k} = a_{k}^{*}-a_{k,k}.
\end{aligned}
\end{equation}
where $a_{k}^{*}$ denotes the accuracy on task k of model trained in the multitask manner and $a_{k,k}$ denotes the  accuracy on k-th task when trained up tp task k in C\&F method. To calculate the Intransigence of a method as a whole, we average the values corresponding to each task. So the lower the Intransigence, the better the learning ability of the model.

We follow [4] to measure Forgetting $\mathcal{F}_{k}$  after trained on task k as follows:
\begin{equation}
\begin{aligned}
&\mathcal{F}_{k} = \mathbb{E}_{j=2...t}f_{j}^{k},\\
&f_{j}^{k}= \max_{l\in \left \{  1...k-1\right \} } a_{l,j} - a_{k,j}.
\end{aligned}
\end{equation}
where $a_{l,j}$ is the model's accuracy on task j after trained on task l. To obtain a comprehensive understanding of a method's propensity to forget, we compute the average of the 'Forgetting' values across all tasks. Consequently, a lower 'Forgetting' value is desirable, indicating that the model retains its knowledge from previous tasks more effectively and suffers less from catastrophic forgetting.

\end{document}